\pdfoutput=1
\documentclass[11pt, letterpaper, logo, onecolumn, copyright, numbering]{paper_improved}
\usepackage{booktabs}
\usepackage{enumitem}
\usepackage{array}
\usepackage{colortbl}
\usepackage{xcolor}
\usepackage{caption}
\usepackage{tabularray}
\usepackage[font=small,labelfont=bf,justification=centering]{caption}
\usepackage[utf8]{inputenc} 
\usepackage[T1]{fontenc}    
\usepackage{hyperref}       
\usepackage{url}            
\usepackage{booktabs}       
\usepackage{amsfonts}       
\usepackage{nicefrac}       
\usepackage{microtype}      
\usepackage{xcolor}         
\usepackage{graphicx}

\usepackage{natbib}
\usepackage{verbatim}
\usepackage{inconsolata}
\usepackage{tabularx}
\usepackage{float}
\usepackage{algorithmic}
\usepackage{multirow}
\usepackage{listings}
\usepackage{xcolor}
\usepackage{subfigure}
\usepackage{subcaption}
\usepackage{amsmath}
\usepackage{amssymb}
\usepackage{array}
\usepackage[ruled,linesnumbered]{algorithm2e}
\usepackage[T1]{fontenc}
\usepackage{fancyhdr}

\usepackage[most]{tcolorbox}
\usepackage{alltt}
\usepackage{lineno}
\newtcolorbox{AIbox}[2][]{aibox,title=#2,#1}
\tcbset{
  aibox/.style={
    top=10pt,
    colframe=black,
    colbacktitle=black,
    enhanced,
    center,
    attach boxed title to top left={yshift=-0.1in,xshift=0.15in},
    boxed title style={boxrule=0pt,colframe=white,},
  }
}

\definecolor{lightgray}{rgb}{0.95, 0.95, 0.95}
\definecolor{darkgray}{rgb}{0.4, 0.4, 0.4}
\definecolor{backcolour}{rgb}{0.95,0.95,0.92}
\definecolor{myblue}{rgb}{0.2, 0.4, 0.8} 
\definecolor{mygreen}{rgb}{0.2, 0.6, 0.2} 

\usepackage{amsmath,amssymb,amsfonts}

\let\cite\citep

\title{OmniVLM: A Token-Compressed, Sub-Billion-Parameter Vision-Language Model for Efficient On-Device Inference}

\reportnumber{} 

\author[*,1]{Wei Chen{$^*$}, Zhiyuan Li, Shuo Xin\\~\\ \bf Nexa AI}

\begin{abstract}
We present OmniVLM, a sub-billion-parameter vision-language model for efficient on-device inference. OmniVLM introduces a token compression mechanism that shortens the visual token sequence from 729 to 81 tokens, significantly reducing computational overhead while preserving visual-semantic fidelity. Through a multi-stage training pipeline of pretraining, supervised fine-tuning, and minimal-edit Direct Preference Optimization (DPO), OmniVLM matches the performance of larger models. On multiple benchmarks, including ScienceQA, POPE, and MMMU, OmniVLM outperforms existing baselines such as nanoLLAVA within a 968M-parameter footprint. Empirical results on the same laptop demonstrate a 9.1$\times$ faster time-to-first-token (0.75s vs.\ 6.82s) and a 1.5$\times$ higher decoding speed (29.41 vs.\ 19.20 tokens/s) compared to nanoLLAVA, enabling efficient deployment on edge devices. The model weights are available on Hugging Face at \url{https://huggingface.co/NexaAIDev/OmniVLM-968M}, and inference examples can be found in Appendix \ref{infer_eg}.
\end{abstract}

\footnotetext[2]{Email: alexchen@nexa.ai}

\begin{document}
\maketitle

\section{Introduction}
Vision-language models (VLMs) are vital to on-device artificial intelligence applications, encompassing tasks from basic image captioning and optical character recognition to advanced capabilities such as visual question answering, video comprehension, and user interface analysis. Deploying these models necessitates the effective integration of visual and textual processing while meeting the strict computational and memory limitations of edge devices such as smartphones, laptops, and embedded systems. Industry applications of VLMs have become abundant in recent years: MultiON uses a VLM for result supervision in its LLM agent Q~\cite{multion}; Meta incorporated a VLM in its newly released Orion AR glasses~\cite{metaAR}; Apple developed Ferret-UI \cite{li2024ferret,you2025ferret} based on multimodal LLMs; and Tencent uses a multimodal LLM for its smartphone agent, AppAgent~\cite{zhang2023appagent}. The Qwen team also recently developed Qwen2-VL as a general-purpose VLM agent \cite{wang2024qwen2}.

The productionization of VLMs for edge deployment presents three primary challenges. First, visual input tokenization introduces substantial computational overhead. According to OpenAI's analysis~\cite{openai2023calculate}, processing a single $1024 \times 1024$ pixel image requires 765 tokens, comprising 170 tokens per $512 \times 512$ tile plus an additional 85 tokens of base processing overhead. Second, power consumption poses a critical constraint for deployment on energy-limited devices. Recent empirical studies from Meta reveal that a 7B-parameter model consumes approximately 0.7J per token~\cite{liu2024mobilellmoptimizingsubbillionparameter}. Consequently, processing a $1024 \times 1024$ pixel image (765 tokens) requires approximately 536 J—consuming over 1\% of an iPhone's battery capacity (50 kJ)—before accounting for text processing. Third, existing VLMs with fewer than 2B parameters demonstrate limited visual comprehension capabilities: the 1B-parameter nanoLLAVA achieves only 28.6\% accuracy on the MMMU benchmark~\cite{nguyen2024nanollava}, significantly underperforming OpenAI's o1 model at 92.3\%~\cite{openai2024simpleevals}. These challenges underscore the complexity of optimizing the trade-off between computational efficiency, energy consumption, and model performance for practical edge deployments.

\newpage 

To address these limitations, we present OmniVLM, a sub-billion-parameter (968M) multimodal model specifically optimized for edge-device deployment. Our key contributions include: (1) a novel token compression mechanism that reduces image token requirements by $9\times$ while maintaining visual fidelity, and (2) an enhanced output-quality framework utilizing Direct Preference Optimization (DPO) with minimal parameter updates. Through comprehensive evaluation on established benchmarks including MM-VET and ScienceQA, we demonstrate that OmniVLM achieves competitive performance while maintaining deployment feasibility on resource-constrained devices. Our research advances the state-of-the-art in efficient vision-language modeling, enabling practical deployment of multimodal capabilities across a broad range of edge computing scenarios.

\section{Related Work}
\subsection{Vision-Language Models: Evolution and Architectures}

Building instruction-following agents has been an active field of study in computer vision. Vision-language models have achieved impressive performance since the introduction of LLaVA~\cite{liu2023llava,liu2023improvedllava}, a large multimodal model that combines a vision encoder and a language model to create an end-to-end trained system for general-purpose visual and language understanding. That work generated multimodal instruction-following data using GPT-4, thereby addressing the scarcity of such data, and introduced two evaluation benchmarks with diverse and challenging tasks to assess the model's capabilities. Additionally, when fine-tuned on ScienceQA~\cite{lu2022learn}, a multimodal reasoning dataset, LLaVA in conjunction with GPT-4 achieves state-of-the-art accuracy. Research on vision-language models has been further advanced by efforts in both academia and industry. Recent progress includes PaliGemma~\cite{beyer2024paligemma} built on top of Gemma-2B~\cite{team2024gemma}, Qwen2-VL~\cite{Qwen-VL,Qwen2VL} built on top of QwenLM~\cite{qwen2}, Llama3.2-Vision built on top of Llama~\cite{llama3-2,dubey2024llama}, and InternVL~\cite{chen2024internvl} built from a vision transformer~\cite{dosovitskiy2020image_ViT} with language models such as QLLaMA~\cite{cui2023efficient_Qllama} and Vicuna~\cite{zheng2023judging_vicuna} as middleware.

\subsection{Edge Deployment and Inference Optimization} 

Due to constraints on memory and computing capability, deploying language models on edge devices such as smartphones and PCs has been a challenge. Techniques such as quantization, pruning, and KV caching are widely used. The GGML library~\cite{ggml} provides the necessary functions for fast inference of large language models, along with the GGUF format for model weights, on top of which abundant high-level inference frameworks have been developed, such as llama.cpp~\cite{llama.cpp} and ctransformers~\cite{ctransformers}. Applications \cite{chen2024octopus, chen2024squid} designed specifically for local execution have emerged in recent years, with notable examples including Ollama~\cite{ollama} and MLC~\cite{mlc-llm}.

\section{Methodology}

\subsection{Model Architecture Overview}
\begin{figure}[ht]
    \centering
    \includegraphics[width=0.8\textwidth]{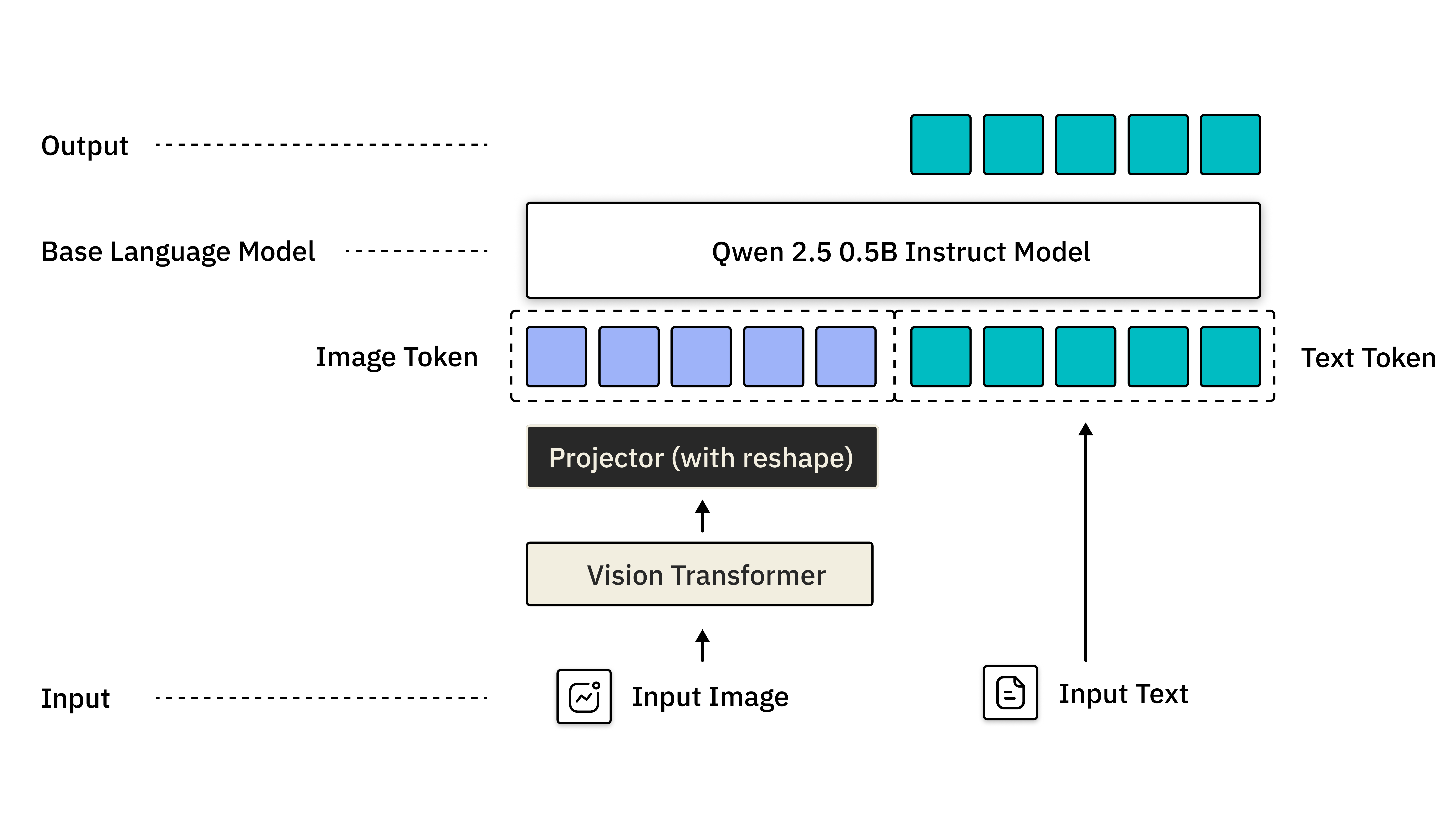}
    \caption{The OmniVLM model architecture.}
    \label{fig:arc}
\end{figure}

The OmniVLM architecture expands on the standard LLaVA architecture \cite{liu2023llava, liu2023improvedllava}. As depicted in Figure \ref{fig:arc}, the vision encoder converts images into embeddings, which are then adjusted by the projection layer to integrate seamlessly with the language model. Compared to the standard LLaVA design, our enhancements reduce computational overhead while maintaining performance.

\begin{itemize}
    \item \textbf{Base Language Model:} The Qwen2.5-0.5B-Instruct model processes text inputs with strong contextual understanding.
    \item \textbf{Vision Encoder:} Google's SigLIP-400M generates high-quality image embeddings at a resolution of $384 \times 384$ with a $14 \times 14$ patch size.
    \item \textbf{Projection Layer:} A Multi-Layer Perceptron (MLP) aligns image embeddings to the token space of the base model. Our novel projection layer reduces the number of image tokens by $9\times$, addressing the inefficiencies in the standard LLaVA architecture.
\end{itemize}

\subsection{Image Token Compression}
Processing image tokens in multimodal models presents significant computational challenges, particularly regarding inference latency and memory utilization. Our LLaVA architecture employs Google's SigLIP-384 module~\cite{zhai2023sigmoidlosslanguageimage}, which generates $729$ tokens per image from a $27 \times 27$ spatial grid, creating a substantial computational bottleneck for real-time applications. To address this limitation, we propose an efficient token compression strategy implemented during the projection phase. This approach transforms the embedding dimensions from $\texttt{[batch\_size}, 729, \texttt{hidden\_size}]$ to $\texttt{[batch\_size}, 81, \texttt{hidden\_size}]$, achieving a 9-fold reduction in token count while maintaining model performance. The compression enables more efficient sequence processing in the language model component.

To optimize the compression methodology, we conducted a comparative analysis of three distinct dimensional transformation techniques: reshaping and two convolution-based approaches. The first convolution variant implements a 1D architecture with kernel size $k=9$ and stride $s=9$ (no padding), while the second employs a 2D configuration with kernel size $(9, 1)$ and corresponding stride $(9, 1)$. Both convolution approaches were designed to reduce sequence dimensionality while preserving spatial information. However, our experimental results demonstrate that these convolution-based methods consistently produce higher validation loss/perplexity than the reshaping strategy.

Based on these empirical findings, we conducted an extensive evaluation of the reshaping approach across four target sequence lengths: 729 (original), 243, 81, and 9. By comparing validation loss/perplexity, we determined that the configuration $\texttt{[batch\_size}, 81, \texttt{hidden\_size}]$ achieves optimal performance, establishing an effective balance between computational efficiency and model capability.

\subsection{Minimal-Edit DPO for Enhanced Response Quality}
Direct Preference Optimization (DPO) has proven highly effective in Reinforcement Learning from Human Feedback (RLHF), significantly reducing hallucinations, improving response correctness, and enhancing model safety. We developed a minimal-edit DPO method in which a teacher model applies small, targeted edits to the base model's outputs. These minimally altered pairs focus on precise improvements in response quality while preserving the original model's behavior. This approach allows for more controlled optimization, securing the benefits of DPO without compromising the stability of the base model. Examples from the DPO pair dataset are presented in Appendix \ref{app_DPO}.

\subsection{Multi-Stage Training}
We developed the \textbf{OmniVLM} model through a three-stage training pipeline:

\begin{itemize}
    \item \textbf{Pretraining:} The model is initialized with foundational visual-linguistic alignments using large-scale image-caption datasets. During this phase, only the projection layer is trained, allowing the model to efficiently learn the mapping between the visual and textual domains. The primary objective is to establish strong visual-linguistic alignments through caption generation tasks, enabling the model to form basic associations between visual features and textual descriptions.

    \item \textbf{Supervised Fine-Tuning (SFT):} The model is fine-tuned on structured datasets comprising image-based question-answer pairs. This stage is designed to enhance the model's contextual understanding, conversational coherence, and response generation capabilities across diverse scenarios. We freeze the vision encoder and train both the projector and the LLM component.

    \item \textbf{Direct Preference Optimization (DPO):} The final stage employs minimal-edit DPO, in which a teacher model corrects the base model's outputs with minor edits. These corrections focus on accuracy-critical elements, forming chosen-rejected pairs for effective preference learning. This stage involves several key steps:
        \begin{itemize}
            \item Deployment of a specialized teacher model to generate minimally edited corrections that preserve essential content while improving quality.
            \item Creation of chosen-rejected pairs by systematically comparing original outputs with their corrected versions.
            \item Application of DPO fine-tuning to align the model's behavior with preferred outputs, effectively teaching it to generate higher-quality responses that match human preferences. In this stage, we keep the vision encoder frozen and train both the projector and the LLM backbone.
        \end{itemize}
\end{itemize}

\section{Experiments}

\subsection{Training Dataset}
We trained OmniVLM through a carefully structured three-stage process, leveraging diverse datasets to achieve robust multimodal capabilities:

\begin{itemize}
    \item \textbf{Pretraining:} In this foundational stage, we utilized large-scale image-caption pairs, primarily sourced from the LLaVA pretraining dataset \cite{liuhaotian2023llavapretrain}. The training focused exclusively on optimizing the connector module, processing approximately 558K training samples.

    \item \textbf{Supervised Fine-Tuning (SFT):} During this critical second stage, we expanded our focus to optimize the entire system architecture. The training incorporated 6M carefully curated samples from the LLaVA dataset \cite{liuhaotian2023llavapretrain}, the UnimmChat dataset \cite{yirany2024unimmchat}, and an internal dataset from Nexa AI.

    \item \textbf{Direct Preference Optimization (DPO):} In the final stage, we leveraged image and prompt data from the RLAIF-V project \cite{openbmb2024rlaif} with chosen and rejected pairs, where the rejected text responses were generated by the model before DPO.
\end{itemize}

\subsection{Compression Ratio Analysis}
We conducted a systematic evaluation of different token compression ratios to determine the optimal balance between computational efficiency and model performance. Starting from the original 729 tokens generated by the SigLIP-384 module, we experimented with four compression targets: 729 (baseline), 243 ($3\times$ reduction), 81 ($9\times$ reduction), and 9 ($81\times$ reduction).

\begin{figure}
    \centering
    \includegraphics[width=0.75\linewidth]{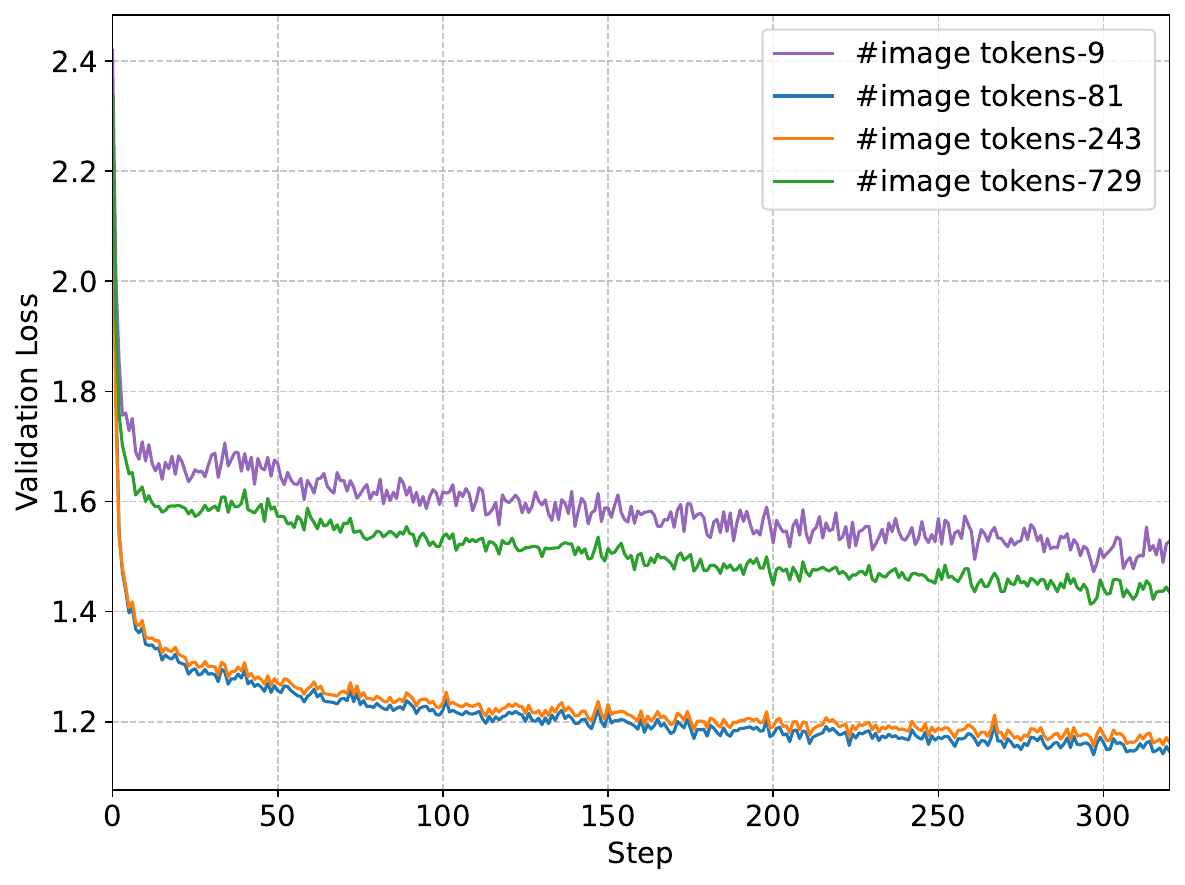}
    \caption{Validation loss curves across different image token compression ratios, evaluated on a test dataset comprising approximately 500K text-image pairs. The comparison demonstrates the effect of token reduction from the baseline (729 tokens) to various compression levels (243, 81, and 9 tokens).}
    \label{fig:loss}
\end{figure}

The validation loss curves for different compression ratios are shown in Figure \ref{fig:loss}. Our experiments revealed that the 81-token configuration ($9\times$ compression) achieves the best performance among all tested ratios. This finding aligns with the architectural constraints of our small-scale language model backbone, where excessive token sequences can degrade performance due to attention mechanism limitations and context window constraints.

Our analysis showed that the 729-token baseline exhibited higher validation losses, likely due to the computational burden on the attention mechanisms. While the 243-token configuration showed improved performance, it still retained unnecessary redundancy in the visual representation. The 81-token setting achieved an optimal balance, preserving essential visual information while minimizing computational overhead. In contrast, the 9-token configuration proved too aggressive, resulting in significant information loss and degraded performance across our evaluation metrics.

These results demonstrate that moderate token compression not only reduces computational requirements but can actually enhance model performance by preventing attention-mechanism saturation in resource-constrained architectures. The $9\times$ compression ratio (81 tokens) emerged as the sweet spot for our model, offering substantial efficiency gains without compromising visual understanding capabilities. This finding suggests that for compact vision-language models, aggressive token compression up to a certain threshold can be beneficial for both computational efficiency and model performance.

\subsection{Quality Benchmark}
\begin{figure}[ht]
    \centering
    \includegraphics[width=0.8\textwidth]{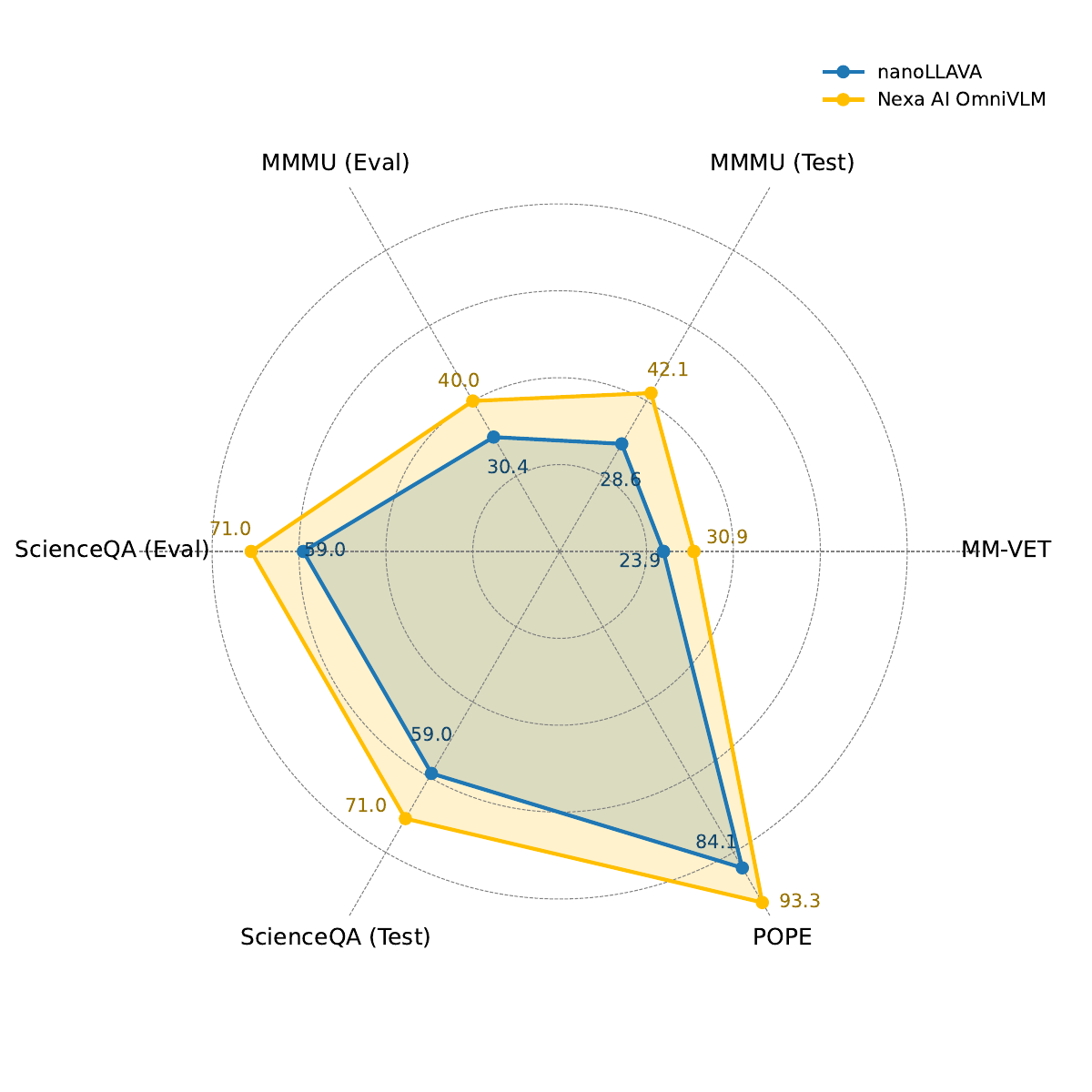}
    \caption{Benchmark comparison of OmniVLM and nanoLLAVA.}
    \label{fig:benchmark}
\end{figure}

In Table \ref{tab:benchmark-results}, we present how OmniVLM outperforms nanoLLAVA, the previous sub-billion-parameter vision-language model in the open-source community. OmniVLM consistently achieves higher scores than nanoLLAVA, showcasing its enhanced reasoning capabilities, multimodal comprehension, and generalization across diverse tasks, as shown in Figure \ref{fig:benchmark}.

Specifically, on the ScienceQA benchmark, OmniVLM achieves a significant improvement, with evaluation and test scores of 71.0, compared to nanoLLAVA’s 59.0. This underscores OmniVLM's ability to better understand and reason about complex science questions that require multimodal input processing. On POPE, OmniVLM exhibits a remarkable gain, achieving 93.3 compared to nanoLLAVA’s 84.1, reflecting its superior performance in precision-oriented tasks. On MM-VET, which evaluates multimodal visual and textual understanding, OmniVLM scores 30.9, outperforming nanoLLAVA by 7.0 points. Similarly, on the MMMU test and evaluation benchmarks, OmniVLM surpasses nanoLLAVA by 13.5 and 9.6 points, respectively, demonstrating its strong multimodal reasoning and contextual integration.

\begin{table}[t]
\centering
\begin{tabular}{lcc}
\toprule
\textbf{Benchmark} & \textbf{OmniVLM} & \textbf{nanoLLAVA} \\
\midrule
ScienceQA (Eval)  & 71.0 & 59.0 \\
ScienceQA (Test)  & 71.0 & 59.0 \\
POPE              & 93.3 & 84.1 \\
MM-VET            & 30.9 & 23.9 \\
MMMU (Test)       & 42.1 & 28.6 \\
MMMU (Eval)       & 40.0 & 30.4 \\
\bottomrule
\end{tabular}
\caption{Performance comparison of OmniVLM and nanoLLAVA on various benchmarks.}
\label{tab:benchmark-results}
\end{table}

\vspace{1em} 

\subsection{Performance on Edge Devices}

We assessed the deployment efficiency of the model on edge devices by evaluating key metrics during local CPU and GPU inference. These metrics include \textbf{Time-to-First-Token (TTFT)}—the latency to generate the first token—and \textbf{Decoding Speed}—tokens generated per second.

\subsubsection{Laptop Performance: AMD Ryzen AI}

Benchmarks were conducted on the ASUS Zenbook S 16 (UM5606), equipped with the AMD Ryzen™ AI processor, under the following configuration:
\begin{itemize}
    \item \textbf{CPU:} AMD Ryzen™ 9 HX 370 (12 cores, 24 threads, 2.0–5.1 GHz, 36 MB cache).
    \item \textbf{GPU:} AMD Radeon™ 890M.
    \item \textbf{Memory:} 24 GB RAM, 1 TB SSD.
\end{itemize}

\begin{table}[ht]
\centering
\caption{Laptop Performance Metrics}
\label{tab:laptop-metrics}
\begin{tabular}{lcc}
\toprule
\textbf{Metric} & \textbf{OmniVLM} & \textbf{nanoLLAVA} \\
\midrule
TTFT (s) & 0.75 & 6.82 \\
Decoding Speed (tokens/s) & 29.41 & 19.20 \\
\bottomrule
\end{tabular}
\end{table}
Performance analysis reveals significant differences between the OmniVLM and nanoLLAVA models, with OmniVLM demonstrating superior performance on both measured metrics. The time-to-first-token (TTFT) for OmniVLM is notably faster at 0.75 seconds compared to nanoLLAVA's 6.82 seconds, representing approximately a $9\times$ improvement in initial response time. In terms of decoding speed, OmniVLM processes 29.41 tokens per second, while nanoLLAVA achieves 19.20 tokens per second, indicating approximately 53\% higher throughput for OmniVLM. These metrics suggest that OmniVLM offers more efficient real-time performance on the tested hardware configuration, particularly in scenarios where rapid response times are crucial.

\subsubsection{Mobile Performance: Samsung Galaxy S22}

The model's performance was also evaluated on the Samsung Galaxy S22 with the following specifications:
\begin{itemize}
    \item \textbf{CPU:} Octa-core (1x3.00 GHz Cortex-X2 \& 3x2.50 GHz Cortex-A710 \& 4x1.80 GHz Cortex-A510).
    \item \textbf{GPU:} Adreno 730.
    \item \textbf{Memory:} 8 GB RAM, 128 GB storage.
\end{itemize}

\begin{table}[ht]
\centering
\caption{Mobile Performance Metrics}
\label{tab:mobile_perm}
\begin{tabular}{lcc}
\toprule
\textbf{Metric} & \textbf{OmniVLM} & \textbf{nanoLLAVA} \\
\midrule
TTFT (s) & 7.48 & 60.23 \\
Decoding Speed (tokens/s) & 31.88 & 24.33 \\
\bottomrule
\end{tabular}
\end{table}

The results in Table \ref{tab:mobile_perm} highlight the significant efficiency of OmniVLM over nanoLLAVA in both TTFT and decoding speed, demonstrating its suitability for latency-sensitive edge applications. On mobile devices, OmniVLM achieves a TTFT of 7.48 seconds, which is approximately $8\times$ faster than nanoLLAVA's 60.23 seconds. This substantial reduction in initial response time is particularly crucial for mobile applications, where user experience depends heavily on rapid system responsiveness.
In terms of decoding speed, OmniVLM processes 31.88 tokens per second compared to nanoLLAVA's 24.33 tokens per second, representing a 31\% improvement in processing throughput. This enhanced decoding efficiency translates into smoother real-time interactions and better overall performance in resource-constrained mobile environments. The performance gains are particularly noteworthy considering the computational limitations typically associated with mobile hardware, suggesting that OmniVLM's architecture is well optimized for edge deployment scenarios.

\section{Future Work and Conclusion}
In this paper, we presented OmniVLM, a compact and efficient sub-billion-parameter vision-language model designed for resource-constrained edge devices. By leveraging innovative techniques such as advanced image token compression, minimal-edit Direct Preference Optimization (DPO), and a multi-stage training pipeline, OmniVLM achieves state-of-the-art performance while drastically reducing computational and energy overhead. Our benchmarks validate OmniVLM's ability to outperform existing models such as nanoLLAVA on linguistic and visual understanding tasks, while remaining practical for deployment on edge hardware, including laptops and mobile devices.

Looking ahead, we aim to extend OmniVLM’s capabilities by optimizing its inference on NPUs (Neural Processing Units) to unlock greater efficiency and scalability across various edge platforms. This includes collaboration with hardware partners to further reduce latency, power consumption, and memory requirements. Our ongoing research seeks to bridge the gap between high-performance multimodal AI and real-world applications, setting a foundation for the next generation of on-device intelligent systems.

\bibliography{main}

\begin{thebibliography}{36}
\providecommand{\natexlab}[1]{#1}
\providecommand{\url}[1]{\texttt{#1}}
\expandafter\ifx\csname urlstyle\endcsname\relax
  \providecommand{\doi}[1]{doi: #1}\else
  \providecommand{\doi}{doi: \begingroup \urlstyle{rm}\Url}\fi

\bibitem[Bai et~al.(2023)Bai, Bai, Yang, Wang, Tan, Wang, Lin, Zhou, and
  Zhou]{Qwen-VL}
J.~Bai, S.~Bai, S.~Yang, S.~Wang, S.~Tan, P.~Wang, J.~Lin, C.~Zhou, and
  J.~Zhou.
\newblock Qwen-vl: A versatile vision-language model for understanding,
  localization, text reading, and beyond.
\newblock \emph{arXiv preprint arXiv:2308.12966}, 2023.

\bibitem[Beyer et~al.(2024)Beyer, Steiner, Pinto, Kolesnikov, Wang, Salz,
  Neumann, Alabdulmohsin, Tschannen, Bugliarello, et~al.]{beyer2024paligemma}
L.~Beyer, A.~Steiner, A.~S. Pinto, A.~Kolesnikov, X.~Wang, D.~Salz, M.~Neumann,
  I.~Alabdulmohsin, M.~Tschannen, E.~Bugliarello, et~al.
\newblock Paligemma: A versatile 3b vlm for transfer.
\newblock \emph{arXiv preprint arXiv:2407.07726}, 2024.

\bibitem[Chen et~al.(2024{\natexlab{a}})Chen, Chen, Zhang, Chen, Wu, Zhang,
  Chen, Li, Wan, and Wang]{chen2024allava}
G.~H. Chen, S.~Chen, R.~Zhang, J.~Chen, X.~Wu, Z.~Zhang, Z.~Chen, J.~Li,
  X.~Wan, and B.~Wang.
\newblock Allava: Harnessing gpt4v-synthesized data for a lite vision-language
  model, 2024{\natexlab{a}}.

\bibitem[Chen and Li(2024)]{chen2024octopus}
W.~Chen and Z.~Li.
\newblock Octopus v2: On-device language model for super agent.
\newblock \emph{arXiv preprint arXiv:2404.01744}, 2024.

\bibitem[Chen et~al.(2024{\natexlab{b}})Chen, Li, Xin, and Wang]{chen2024squid}
W.~Chen, Z.~Li, S.~Xin, and Y.~Wang.
\newblock Squid: Long context as a new modality for energy-efficient on-device
  language models.
\newblock \emph{arXiv preprint arXiv:2408.15518}, 2024{\natexlab{b}}.

\bibitem[Chen et~al.(2024{\natexlab{c}})Chen, Wu, Wang, Su, Chen, Xing, Zhong,
  Zhang, Zhu, Lu, et~al.]{chen2024internvl}
Z.~Chen, J.~Wu, W.~Wang, W.~Su, G.~Chen, S.~Xing, M.~Zhong, Q.~Zhang, X.~Zhu,
  L.~Lu, et~al.
\newblock Internvl: Scaling up vision foundation models and aligning for
  generic visual-linguistic tasks.
\newblock In \emph{Proceedings of the IEEE/CVF Conference on Computer Vision
  and Pattern Recognition}, pages 24185--24198, 2024{\natexlab{c}}.

\bibitem[Community(2023)]{openai2023calculate}
O.~Community.
\newblock How do i calculate image tokens in gpt-4 vision?
\newblock
  \url{https://community.openai.com/t/how-do-i-calculate-image-tokens-in-gpt4-vision/492318},
  2023.
\newblock Accessed: November 28, 2024.

\bibitem[Cui et~al.(2023)Cui, Yang, and Yao]{cui2023efficient_Qllama}
Y.~Cui, Z.~Yang, and X.~Yao.
\newblock Efficient and effective text encoding for chinese llama and alpaca.
\newblock \emph{arXiv preprint arXiv:2304.08177}, 2023.

\bibitem[Dosovitskiy(2020)]{dosovitskiy2020image_ViT}
A.~Dosovitskiy.
\newblock An image is worth 16x16 words: Transformers for image recognition at
  scale.
\newblock \emph{arXiv preprint arXiv:2010.11929}, 2020.

\bibitem[Dubey et~al.(2024)Dubey, Jauhri, Pandey, Kadian, Al-Dahle, Letman,
  Mathur, Schelten, Yang, Fan, et~al.]{dubey2024llama}
A.~Dubey, A.~Jauhri, A.~Pandey, A.~Kadian, A.~Al-Dahle, A.~Letman, A.~Mathur,
  A.~Schelten, A.~Yang, A.~Fan, et~al.
\newblock The llama 3 herd of models.
\newblock \emph{arXiv preprint arXiv:2407.21783}, 2024.

\bibitem[Gerganov(2023-2024{\natexlab{a}})]{ggml}
G.~Gerganov.
\newblock {ggml}, 2023-2024{\natexlab{a}}.
\newblock URL \url{https://github.com/ggerganov/ggml}.

\bibitem[Gerganov(2023-2024{\natexlab{b}})]{llama.cpp}
G.~Gerganov.
\newblock {llama.cpp}, 2023-2024{\natexlab{b}}.
\newblock URL \url{https://github.com/ggerganov/llama.cpp}.

\bibitem[Li et~al.(2024)Li, You, Zhang, Feng, Agrawal, Li, Moorthy, Nichols,
  Yang, and Gan]{li2024ferret}
Z.~Li, K.~You, H.~Zhang, D.~Feng, H.~Agrawal, X.~Li, M.~P.~S. Moorthy,
  J.~Nichols, Y.~Yang, and Z.~Gan.
\newblock Ferret-ui 2: Mastering universal user interface understanding across
  platforms.
\newblock \emph{arXiv preprint arXiv:2410.18967}, 2024.

\bibitem[Liu(2023)]{liuhaotian2023llavapretrain}
H.~Liu.
\newblock Llava-pretrain dataset.
\newblock \url{https://huggingface.co/datasets/liuhaotian/LLaVA-Pretrain},
  2023.
\newblock Accessed: November 28, 2024.

\bibitem[Liu et~al.(2023{\natexlab{a}})Liu, Li, Li, and
  Lee]{liu2023improvedllava}
H.~Liu, C.~Li, Y.~Li, and Y.~J. Lee.
\newblock Improved baselines with visual instruction tuning,
  2023{\natexlab{a}}.

\bibitem[Liu et~al.(2023{\natexlab{b}})Liu, Li, Wu, and Lee]{liu2023llava}
H.~Liu, C.~Li, Q.~Wu, and Y.~J. Lee.
\newblock Visual instruction tuning.
\newblock In \emph{NeurIPS}, 2023{\natexlab{b}}.

\bibitem[Liu et~al.(2024)Liu, Zhao, Iandola, Lai, Tian, Fedorov, Xiong, Chang,
  Shi, Krishnamoorthi, Lai, and
  Chandra]{liu2024mobilellmoptimizingsubbillionparameter}
Z.~Liu, C.~Zhao, F.~Iandola, C.~Lai, Y.~Tian, I.~Fedorov, Y.~Xiong, E.~Chang,
  Y.~Shi, R.~Krishnamoorthi, L.~Lai, and V.~Chandra.
\newblock Mobilellm: Optimizing sub-billion parameter language models for
  on-device use cases, 2024.
\newblock URL \url{https://arxiv.org/abs/2402.14905}.

\bibitem[Lu et~al.(2022)Lu, Mishra, Xia, Qiu, Chang, Zhu, Tafjord, Clark, and
  Kalyan]{lu2022learn}
P.~Lu, S.~Mishra, T.~Xia, L.~Qiu, K.-W. Chang, S.-C. Zhu, O.~Tafjord, P.~Clark,
  and A.~Kalyan.
\newblock Learn to explain: Multimodal reasoning via thought chains for science
  question answering.
\newblock In \emph{The 36th Conference on Neural Information Processing Systems
  (NeurIPS)}, 2022.

\bibitem[Marella(2023-2024)]{ctransformers}
R.~Marella.
\newblock {ctransformers}, 2023-2024.
\newblock URL \url{https://github.com/marella/ctransformers}.

\bibitem[{Meta}(2024{\natexlab{a}})]{llama3-2}
{Meta}.
\newblock Llama 3.2: Revolutionizing edge ai and vision with open, customizable
  models, 2024{\natexlab{a}}.
\newblock URL
  \url{https://ai.meta.com/blog/llama-3-2-connect-2024-vision-edge-mobile-devices/}.
\newblock Accessed: November 28, 2024.

\bibitem[{Meta}(2024{\natexlab{b}})]{metaAR}
{Meta}.
\newblock Introducing orion, our first true augmented reality glasses,
  2024{\natexlab{b}}.
\newblock URL
  \url{https://about.fb.com/news/2024/09/introducing-orion-our-first-true-augmented-reality-glasses/}.
\newblock Accessed: November 28, 2024.

\bibitem[{MLC team}(2023-2024)]{mlc-llm}
{MLC team}.
\newblock {MLC-LLM}, 2023-2024.
\newblock URL \url{https://github.com/mlc-ai/mlc-llm}.

\bibitem[{MultiON team}(2024)]{multion}
{MultiON team}.
\newblock Introducing agent q: Research breakthrough for the next generation of
  ai agents with planning and self healing capabilities, 2024.
\newblock URL
  \url{https://www.multion.ai/blog/introducing-agent-q-research-breakthrough-for-the-next-generation-of-ai-agents-with-planning-and-self-healing-capabilities}.
\newblock Accessed: November 28, 2024.

\bibitem[Nguyen(2024)]{nguyen2024nanollava}
Q.~Nguyen.
\newblock nanollava: A compact multi-modal model for edge devices.
\newblock \url{https://huggingface.co/qnguyen3/nanoLLaVA}, 2024.
\newblock Accessed: November 28, 2024.

\bibitem[{Ollama team}(2023-2024)]{ollama}
{Ollama team}.
\newblock {Ollama}, 2023-2024.
\newblock URL \url{https://ollama.com/}.

\bibitem[{OpenAI}(2024)]{openai2024simpleevals}
{OpenAI}.
\newblock Simple-evals: Openai benchmark suite for language model evaluation.
\newblock \url{https://github.com/openai/simple-evals}, 2024.
\newblock Accessed: November 28, 2024.

\bibitem[{OpenBMB}(2024)]{openbmb2024rlaif}
{OpenBMB}.
\newblock Rlaif-v dataset.
\newblock \url{https://huggingface.co/datasets/openbmb/RLAIF-V-Dataset}, 2024.
\newblock Accessed: November 28, 2024.

\bibitem[Team et~al.(2024)Team, Mesnard, Hardin, Dadashi, Bhupatiraju, Pathak,
  Sifre, Rivi{\`e}re, Kale, Love, et~al.]{team2024gemma}
G.~Team, T.~Mesnard, C.~Hardin, R.~Dadashi, S.~Bhupatiraju, S.~Pathak,
  L.~Sifre, M.~Rivi{\`e}re, M.~S. Kale, J.~Love, et~al.
\newblock Gemma: Open models based on gemini research and technology.
\newblock \emph{arXiv preprint arXiv:2403.08295}, 2024.

\bibitem[Wang et~al.(2024{\natexlab{a}})Wang, Bai, Tan, Wang, Fan, Bai, Chen,
  Liu, Wang, Ge, Fan, Dang, Du, Ren, Men, Liu, Zhou, Zhou, and Lin]{Qwen2VL}
P.~Wang, S.~Bai, S.~Tan, S.~Wang, Z.~Fan, J.~Bai, K.~Chen, X.~Liu, J.~Wang,
  W.~Ge, Y.~Fan, K.~Dang, M.~Du, X.~Ren, R.~Men, D.~Liu, C.~Zhou, J.~Zhou, and
  J.~Lin.
\newblock Qwen2-vl: Enhancing vision-language model's perception of the world
  at any resolution.
\newblock \emph{arXiv preprint arXiv:2409.12191}, 2024{\natexlab{a}}.

\bibitem[Wang et~al.(2024{\natexlab{b}})Wang, Bai, Tan, Wang, Fan, Bai, Chen,
  Liu, Wang, Ge, et~al.]{wang2024qwen2}
P.~Wang, S.~Bai, S.~Tan, S.~Wang, Z.~Fan, J.~Bai, K.~Chen, X.~Liu, J.~Wang,
  W.~Ge, et~al.
\newblock Qwen2-vl: Enhancing vision-language model's perception of the world
  at any resolution.
\newblock \emph{arXiv preprint arXiv:2409.12191}, 2024{\natexlab{b}}.

\bibitem[Yang et~al.(2024)Yang, Yang, Hui, Zheng, Yu, Zhou, Li, Li, Liu, Huang,
  Dong, Wei, Lin, Tang, Wang, Yang, Tu, Zhang, Ma, Xu, Zhou, Bai, He, Lin,
  Dang, Lu, Chen, Yang, Li, Xue, Ni, Zhang, Wang, Peng, Men, Gao, Lin, Wang,
  Bai, Tan, Zhu, Li, Liu, Ge, Deng, Zhou, Ren, Zhang, Wei, Ren, Fan, Yao,
  Zhang, Wan, Chu, Liu, Cui, Zhang, and Fan]{qwen2}
A.~Yang, B.~Yang, B.~Hui, B.~Zheng, B.~Yu, C.~Zhou, C.~Li, C.~Li, D.~Liu,
  F.~Huang, G.~Dong, H.~Wei, H.~Lin, J.~Tang, J.~Wang, J.~Yang, J.~Tu,
  J.~Zhang, J.~Ma, J.~Xu, J.~Zhou, J.~Bai, J.~He, J.~Lin, K.~Dang, K.~Lu,
  K.~Chen, K.~Yang, M.~Li, M.~Xue, N.~Ni, P.~Zhang, P.~Wang, R.~Peng, R.~Men,
  R.~Gao, R.~Lin, S.~Wang, S.~Bai, S.~Tan, T.~Zhu, T.~Li, T.~Liu, W.~Ge,
  X.~Deng, X.~Zhou, X.~Ren, X.~Zhang, X.~Wei, X.~Ren, Y.~Fan, Y.~Yao, Y.~Zhang,
  Y.~Wan, Y.~Chu, Y.~Liu, Z.~Cui, Z.~Zhang, and Z.~Fan.
\newblock Qwen2 technical report.
\newblock \emph{arXiv preprint arXiv:2407.10671}, 2024.

\bibitem[Yirany(2024)]{yirany2024unimmchat}
Yirany.
\newblock Unimm-chat dataset.
\newblock \url{https://huggingface.co/datasets/Yirany/UniMM-Chat}, 2024.
\newblock Accessed: November 28, 2024.

\bibitem[You et~al.(2025)You, Zhang, Schoop, Weers, Swearngin, Nichols, Yang,
  and Gan]{you2025ferret}
K.~You, H.~Zhang, E.~Schoop, F.~Weers, A.~Swearngin, J.~Nichols, Y.~Yang, and
  Z.~Gan.
\newblock Ferret-ui: Grounded mobile ui understanding with multimodal llms.
\newblock In \emph{European Conference on Computer Vision}, pages 240--255.
  Springer, 2025.

\bibitem[Zhai et~al.(2023)Zhai, Mustafa, Kolesnikov, and
  Beyer]{zhai2023sigmoidlosslanguageimage}
X.~Zhai, B.~Mustafa, A.~Kolesnikov, and L.~Beyer.
\newblock Sigmoid loss for language image pre-training, 2023.
\newblock URL \url{https://arxiv.org/abs/2303.15343}.

\bibitem[Zhang et~al.(2023)Zhang, Yang, Liu, Han, Chen, Huang, Fu, and
  Yu]{zhang2023appagent}
C.~Zhang, Z.~Yang, J.~Liu, Y.~Han, X.~Chen, Z.~Huang, B.~Fu, and G.~Yu.
\newblock Appagent: Multimodal agents as smartphone users.
\newblock \emph{arXiv preprint arXiv:2312.13771}, 2023.

\bibitem[Zheng et~al.(2023)Zheng, Chiang, Sheng, Zhuang, Wu, Zhuang, Lin, Li,
  Li, Xing, et~al.]{zheng2023judging_vicuna}
L.~Zheng, W.-L. Chiang, Y.~Sheng, S.~Zhuang, Z.~Wu, Y.~Zhuang, Z.~Lin, Z.~Li,
  D.~Li, E.~Xing, et~al.
\newblock Judging llm-as-a-judge with mt-bench and chatbot arena.
\newblock \emph{Advances in Neural Information Processing Systems},
  36:\penalty0 46595--46623, 2023.

\end{thebibliography}

\appendix
\section{DPO Dataset Preparation}
\label{app_DPO}
Our dataset preparation process leverages GPT-4V to generate synthetic training pairs with minimal edit distance. We applied this approach to generate a series of datasets for several stages of DPO training. For instance, we used the GPT-4V-generated image caption data in the ALLaVA dataset~\cite{chen2024allava} and produced a DPO dataset through minimal edits to the captions. Table \ref{DPO_example} lists two examples of the generated data.

\begin{table}[h!]
\centering
\begin{tabular}{|p{1.5cm}|p{7cm}|p{7cm}|}
\hline
\textbf{prompt}  & Please provide a detailed narrative of the image. & Please illustrate the image through your words. \\ \hline
 \textbf{image} & 
\includegraphics[height=3cm]{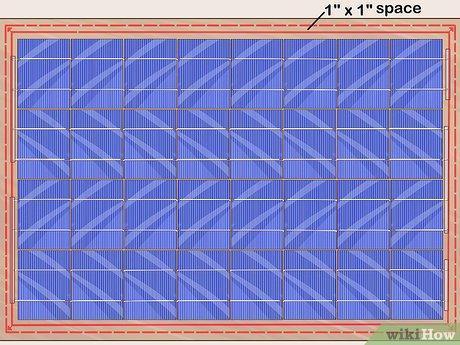} & 
\includegraphics[height=3cm]{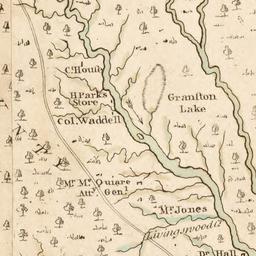} \\ \hline
\textbf{chosen}   &  \small The image displays a schematic representation of a quilted grid pattern with blue diagonal lines on a white background, symbolizing fabric, likely silk, due to the context of the question. Each grid section is separated by thin white lines, and a red-orange double border frames the entire grid. The top right corner has text indicating a "1" x 1" space", suggesting measurements for the grid. There's no physical garment or ironing tools visible, which implies that the image is a conceptual or instructional illustration rather than a photograph of the actual task.   & \small The image is a section of an old map, presented in sepia tones indicative of aged parchment. It features handwritten labels, stylized iconography for trees, and several delineated plots of land, typical of historical land grant maps. Notable names are marked alongside illustrations of houses, perhaps indicating property ownership. A prominent river snakes through the center with tributary streams, and the scale suggests a close-up of a rural area. The absence of grid lines or latitude and longitude marks suggests it predates modern cartographic conventions.  \\ \hline
\textbf{rejected}   & \small The image displays a diagram or illustration of a grid of blue and white squares, arranged in a pattern that resembles a solar panel array. The squares are arranged in a 3x3 grid, with each square containing a blue color and white lines. The blue squares are likely representing the photovoltaic cells, while the white lines may indicate the conductive pathways or the edges of the solar panels. The image also contains a red border around the grid, and there is a note indicating a "1" x 1" space, suggesting a specific measurement or spacing between the cells. The overall image appears to be a step-by-step guide or instructional material, possibly related to solar panel installation or maintenance.  & \small The image is a map, depicting a geographic area with various geographic features and locations. The map is detailed, showing topographical elements such as rivers, mountains, and landforms. There are several labeled areas, including "Griffin," "Cottwood," and "M. McQuarrie." The map is colored in shades of green, brown, and beige, with the green representing land and the brown and beige indicating water bodies. The map is not a satellite view, but rather a more detailed and close-up view of the area. There are no people or animals visible in the image.   \\ \hline
\end{tabular}
\caption{Examples of the DPO dataset generated from {allava\_vflan} image caption data with minimal edit distance.}
\label{DPO_example}
\end{table}

\section{Image Inference Examples}\label{infer_eg}
Based on community feedback, we continuously refined our model across five key dimensions:

\begin{itemize}
    \item \textbf{Artistic Description Enhancement}: Improved recognition and description of artistic styles, techniques, and compositional elements
    
    \item \textbf{Complex Scene Analysis}: Advanced capabilities in understanding multi-object relationships and contextual interactions
    
    \item \textbf{Anime/Manga Recognition}: Specialized improvements in identifying anime-specific visual elements and character traits
    
    \item \textbf{Detail Perception}: Enhanced accuracy in color detection, texture analysis, and fine-grained visual feature recognition
    
    \item \textbf{Knowledge Integration}: Expanded contextual understanding through integration of broader world knowledge
\end{itemize}

The following examples demonstrate OmniVLM's capabilities across these dimensions.

\begin{figure}[ht]
    \centering
    \includegraphics[width=0.7\textwidth]{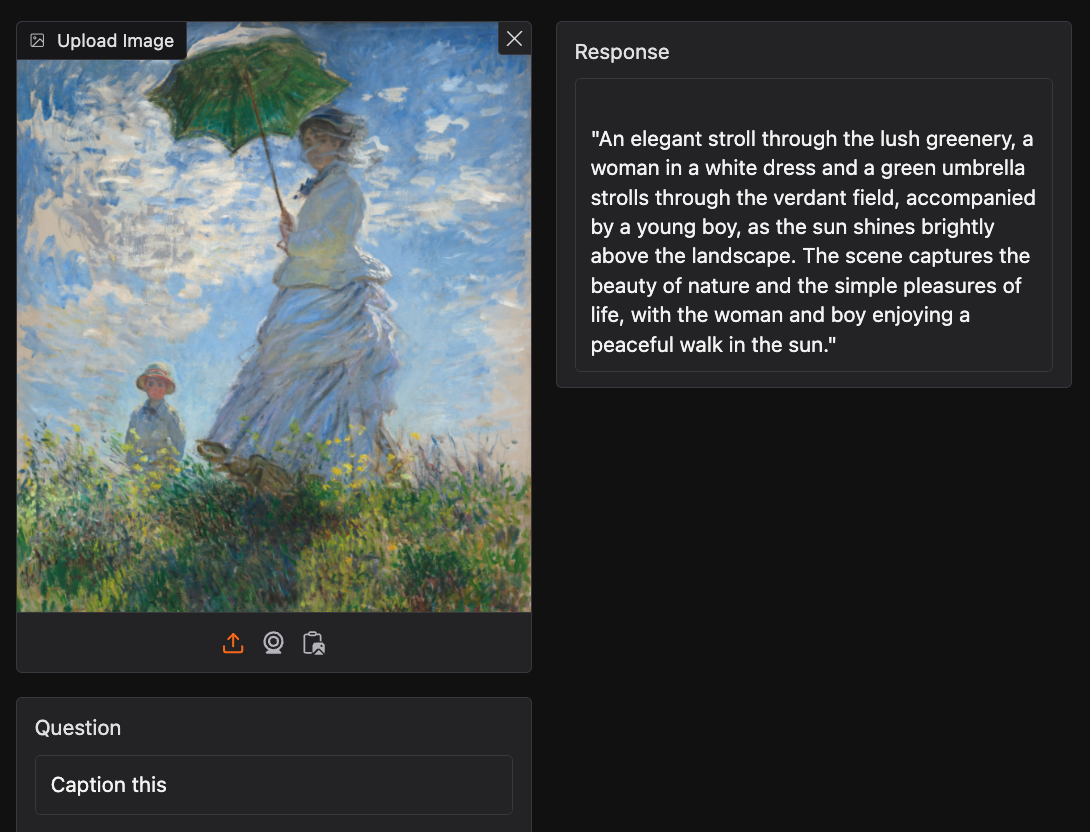}
    \caption{Example of art description.}
    \label{fig:art}
\end{figure}

\begin{figure}[ht]
    \centering
    \includegraphics[width=0.7\textwidth]{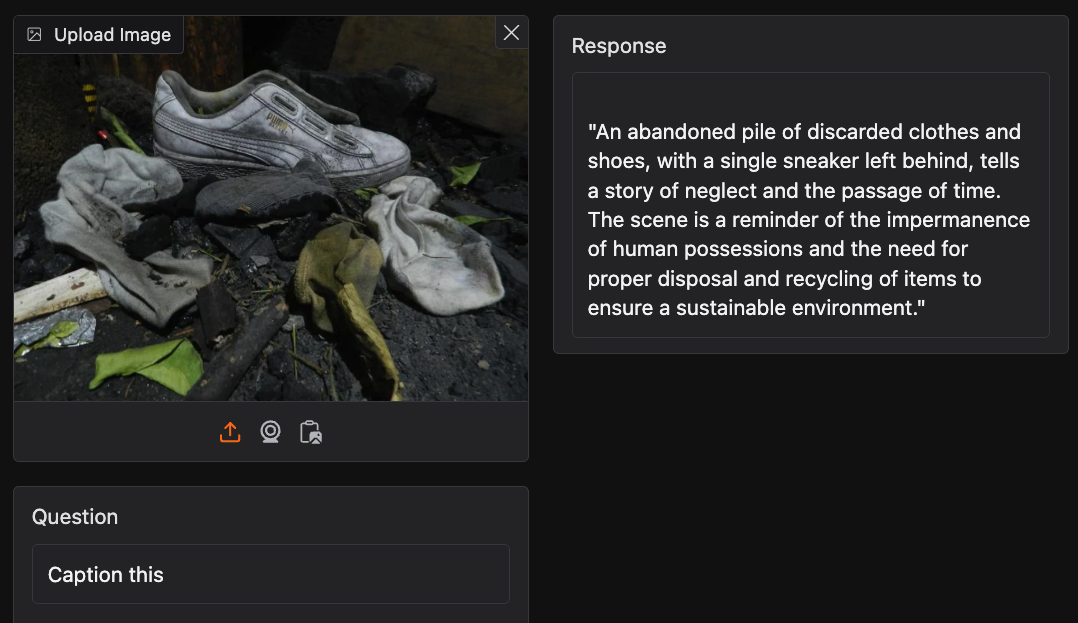}
    \caption{Example of complex scene analysis.}
    \label{fig:complex}
\end{figure}
\begin{figure}[ht]
    \centering
    \includegraphics[width=0.7\textwidth]{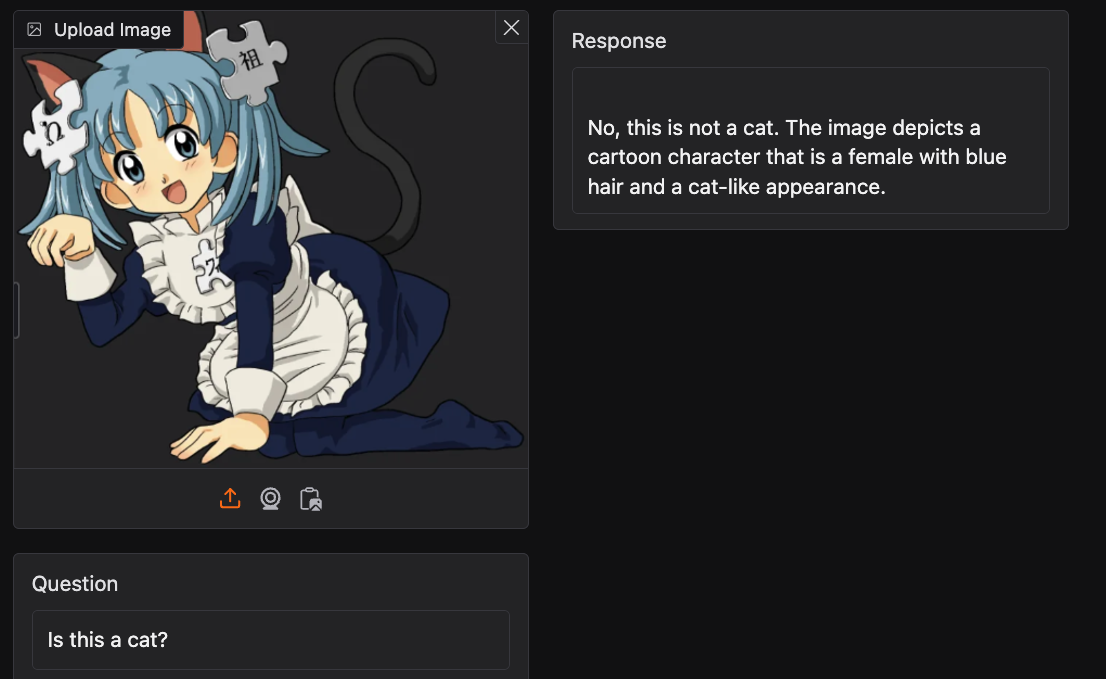}
    \caption{Example of anime recognition.}
    \label{fig:anime}
\end{figure}
\begin{figure}[ht]
    \centering
    \includegraphics[width=0.7\textwidth]{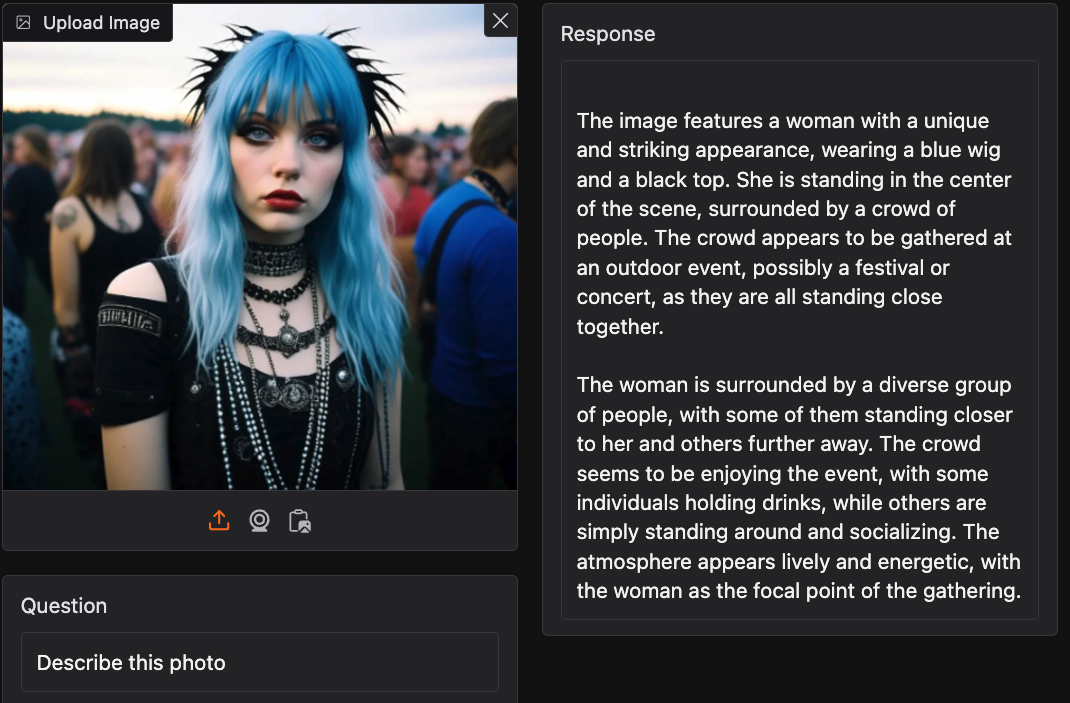}
    \caption{Example of detail perception.}
    \label{fig:detail}
\end{figure}
\begin{figure}[ht]
    \centering
    \includegraphics[width=0.7\textwidth]{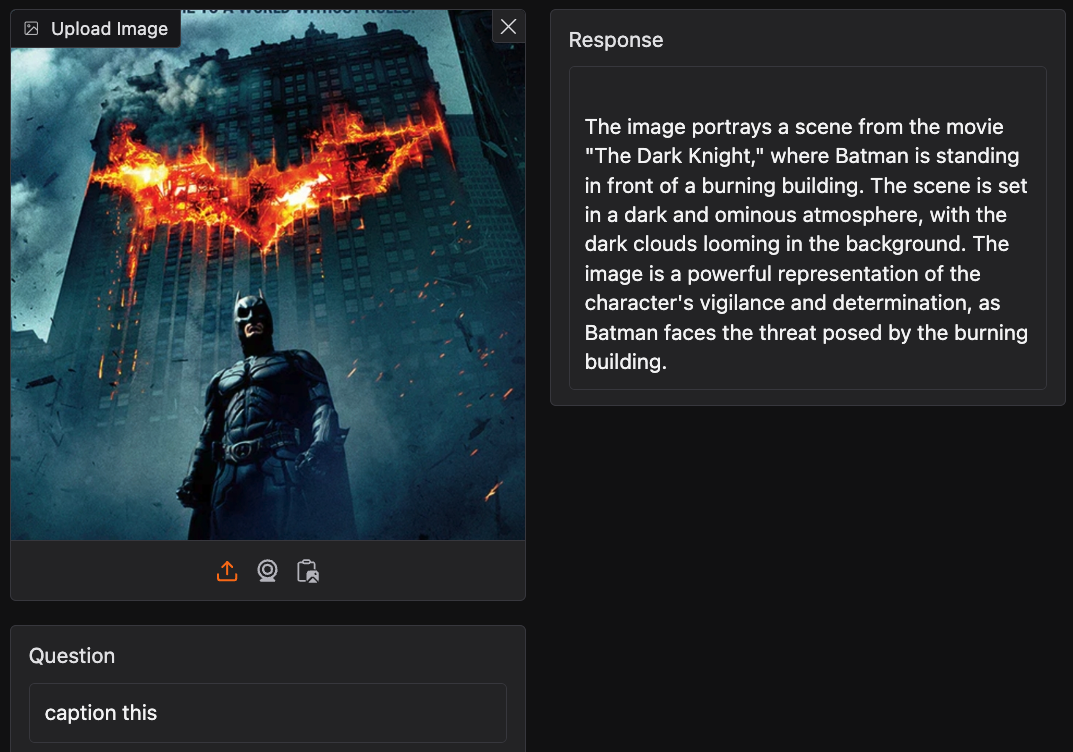}
    \caption{Example of world knowledge.}
    \label{fig:world}
\end{figure}

\end{document}